\documentclass[pdflatex,sn-mathphys-num,iicol]{sn-jnl}

\usepackage{graphicx}%
\usepackage{multirow}%
\usepackage{amsmath,amssymb,amsfonts}%
\usepackage{amsthm}%
\usepackage{mathrsfs}%
\usepackage[title]{appendix}%
\usepackage{xcolor}%
\usepackage{textcomp}%
\usepackage{manyfoot}%
\usepackage{booktabs}%
\usepackage{algorithm}%
\usepackage{algorithmicx}%
\usepackage{algpseudocode}%
\usepackage{listings}%
\usepackage{orcidlink}%

\usepackage{xurl}
\usepackage{placeins}
\usepackage{subfigure}
\usepackage{dblfloatfix}
\usepackage{hyperref}
\usepackage{acronym}
\usepackage{siunitx}
\usepackage{bm}
\usepackage{svg}
\usepackage{newtxmath} 
\usepackage[scr=rsfso]{mathalfa} 
\usepackage[noabbrev, capitalise]{cleveref}
\usepackage[perpage]{footmisc}
\allowdisplaybreaks

\acrodef{RL}{Reinforcement Learning}
\acrodef{NN}{Neural Network}
\acrodef{DoF}{Degree of Freedom}
\acrodef{ID}{Inverse Dynamics}
\acrodef{IMU}{Inertial Measurement Unit}
\acrodef{PPO}{Proximal Policy Optimization}
\acrodef{PSD}{Power Spectral Density}
\acrodef{MLP}{Multi-Layer Perceptron}
\acrodef{COG}{Center Of Gravity}
\acrodef{EE}{end effector}
\acrodef{AOA}{angle of arrival}
\acrodef{RF}{radio frequency}
\acrodef{SDR}{Software Defined Radio}
\acrodef{SM}{Switching Matrix}
\acrodef{CW}{Continuous Wave}
\acrodef{ISM}{Industrial, Scientific, and Medical}
\acrodef{TDM}{Time Devision Multiplexing}

\newcommand{\eqHigh}[1]{\colorbox{lightgray}{$\displaystyle #1$}}

\theoremstyle{thmstyleone}%
\theoremstyle{thmstyletwo}%

\theoremstyle{thmstylethree}%

\begin{document}

\title[Calibrated Dynamic Modeling for Force and Payload Estimation in Hydraulic Machinery]{Calibrated Dynamic Modeling for Force and Payload Estimation in Hydraulic Machinery}


\author*[1]{\fnm{Lennart} \sur{Werner}\orcidlink{0000-0002-1338-0458}}\email{lennartwerner@ethz.ch}
\author[1]{\fnm{Pol} \sur{Eyschen}\orcidlink{0009-0001-2371-4116}}\email{peyschen@ethz.ch}

\author[2]{\fnm{Sean} \sur{Costello}}\email{sean.costello@hexagon.com}
\author[2]{\fnm{Pierluigi} \sur{Micarelli}}\email{pierluigi.micarelli@hexagon.com}

\author[1]{\fnm{Marco} \sur{Hutter}\orcidlink{0000-0002-4285-4990}}\email{mahutter@ethz.ch}

\affil*[1]{\orgdiv{ETH Zürich}, \orgname{Robotic Systems Lab}, \orgaddress{\street{Leonhardstrasse 21}, \city{Zürich}, \postcode{8092}, \country{Switzerland}}}

\affil[2]{\orgdiv{Hexagon AB}, \orgaddress{\street{Heinrich-Wild-Strasse 201}, \city{Heerbrugg}, \postcode{9435}, \country{Switzerland}}}


\abstract{Accurate real-time estimation of end effector interaction forces in hydraulic excavators is a key enabler for advanced automation in heavy machinery.
Accurate knowledge of these forces allows improved, precise grading and digging manoeuvers.
To address these challenges, we introduce a high-accuracy, retrofittable 2D force- and payload estimation algorithm that does not impose additional requirements on the operator regarding trajectory, acceleration or the use of the slew joint. 
The approach is designed for retrofittability, requires minimal calibration and no prior knowledge of machine-specific dynamic characteristics.
Specifically, we propose a method for identifying a dynamic model, necessary to estimate both end effector interaction forces and bucket payload during normal operation.
Our optimization-based payload estimation achieves a full-scale payload accuracy of 1\%.
On a standard \SI{25}{t} excavator, the online force measurement from pressure and inertial measurements achieves a direction accuracy of \SI{13}{\degree} and a magnitude accuracy of \SI{383}{\N}.
The method's accuracy and generalization capability are validated on two excavator platforms of different type and weight classes.
We benchmark our payload estimation against a classical quasistatic method and a commercially available system.
Our system outperforms both in accuracy and precision.
}




\maketitle

\section{Introduction}\label{sec1}
\let\thefootnote\relax\footnotetext{This project received funding from Hexagon AB}
The traditionally labor-intensive construction sector is gaining attention from the automation research community~\cite{cai2019construction}. 
Automation of heavy construction machinery aims to reduce workplace accidents, worker fatigue, address labor shortages, and improve precision and safety on construction sites.
In particular, assistive functions can help less skilled workers perform tasks that usually require experienced operators, and developing these modes is the first step toward full automation.
Earthwork tasks such as excavation and grading introduce additional complexity due to the interaction between the tool and the soil.

Different soil types require different excavation methods: soft or dry soil can be dug with deep scoops, while dense or compact soil requires scraping motions. 
Despite advances in learning-based techniques for trajectory adaption to soil properties~\cite{Egli22SoilAdaptiveExcavation}, 
subsurface obstacles such as boulders and pipes remain a challenge.
The use of interaction forces between the tool and the soil has the potential to improve control and ease of handling buried obstacles.

This is why we investigate a precise method for estimating in-plane \ac{EE} interaction forces and bucket payloads in hydraulic construction machinery. 
Across literature, force estimation is achieved by physically modeling the soil~\cite{Zhao_Wang_Zhang_Luo_2020}, fitting data-driven models to sensor readings~\cite{Azure_Ayawah_Kaba_Kadingdi_Frimpong_2021} or explicitly modeling the system dynamics~\cite{ferlibas2021load}.
We adopt the approach of explicitly modeling the system dynamics due to the high accuracy and 
potential for generalization across different motions and terrains.
Our method is designed to be retrofittable, requiring minimal machine-specific calibration and no prior knowledge of inertial parameters. 
We only require kinematic measurements of the arm as well as pressure sensors on two arm cylinders.
Our technique works without imposing motion constraints on the operator, such as smooth trajectories or a specific weighing workspace.

During deployment, our general, parametrized dynamics model predicts zero-load system torques purely from joint position, velocity and acceleration.
By comparing these predicted system torques with those measured from cylinder pressure readings, our two independent methods can estimate end-effector force magnitude, direction, and bucket payload.
While our \ac{EE} force vector computation outputs live data, the payload estimation records a trajectory and uses optimization in order to predict a 1\% full-scale accurate weight estimate of the bucket payload.
Both estimators are computationally lightweight and can be run on embedded hardware during deployment.

\begin{figure}
    \centering
    \includegraphics[width=\linewidth]{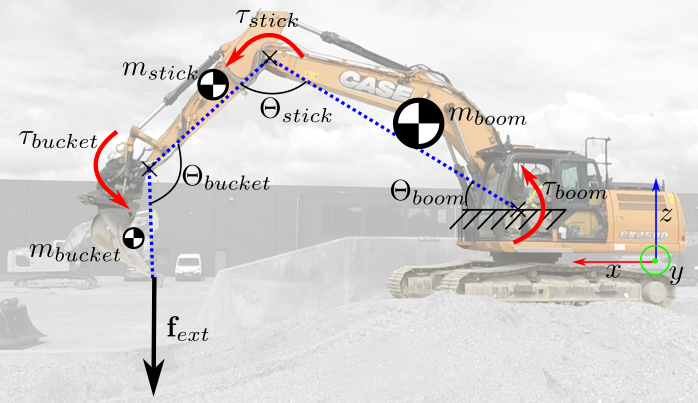}
    \caption{CASE250 excavator used for evaluation. Definition and naming of angles, torques and external force.}
    \label{fig:naming}
\end{figure}

The performance and accuracy of the force and payload estimation is dependent on the correct estimation of the free parameters in the parameterized dynamic model.
Because the correct estimation of parameters is crucial for the performance and accuracy, specific emphasis of this work lies on the calibration of these parameters.
We present a novel calibration routine that enables this work to be adopted for diverse hydraulic excavators in a retrofitting process.
In a benchmark on two different full-sized machines, we are demonstrating generalizability and accuracy of our method.
We compare the performance of our method against a classical quasi-static payload estimation~\cite{bennet2014payload}, which works with a lookup function from joint angles to system torques. 
As a second reference, we also compare the results with a commercially available system with unknown technique.
Over the same 55 loading cycles on a full-scale excavator as shown in~\cref{fig:naming}, the classical quasi-static payload estimation algorithm yields a mean error of \SI{264.7}{\kg} and a standard deviation of \SI{2263}{\kg}, while our method achieves a significantly lower mean error of \SI{5.7}{\kg} and a standard deviation of \SI{30.4}{\kg}.%
In addition, our system outperforms the commercial payload estimation solution with an average error of \SI{101.7}{\kg} and a standard deviation of \SI{51.3}{\kg}.%

In sum, the contributions of this paper are:
\begin{itemize}
    \item A high accuracy method to predict system torques in all operating conditions without additional constraints on the motion regarding trajectory, acceleration, or the use of the slew joint.
    \item A novel calibration routine to identify essential dynamics parameters on existing machinery without the need for disassembly.
    \item A computationally lightweight end effector force and 1\% full-scale accuracy bucket payload estimator using the predicted system torques.
\end{itemize}

\section{Related Work}
\label{sec:related_work}
\subsection{Payload- and Force Estimation}
Online bucket payload measurement systems are well documented in research and widely available as commercial solutions \cite{CATPayload,trimble,topcon}.
The Patent by Silvy-Leligois and Giroud~\cite{pat1} describes a technique based on the empirical identification of static torques.
Two calibrated payloads are used with a predefined motion for model identification. 
Inertia effects, as well as dynamic torques, remain unconsidered, constraining the range of motions usable by the technique~\cite{bennet2014payload}.
Renner et al.~\cite{renner2020online} propose a generalized online payload estimation method for hydraulic manipulators.
The work formulates the manipulator dynamics as a linear regression model, with adjustments for cylinder friction and compensations for closed kinematic chains.
The authors then define a least squares optimization problem in order to solve for the payload parameters.
In a post-processing step, a weighted mean is applied over the estimated payload mass to deal with spikes during loading and unloading.
However, in order to parametrize the dynamic model, extensive prior knowledge of the system is needed, such as inertia and masses of individual links.
We propose to forgo any prior information and introduce a calibration routine that allows our method to be retrofitted to any suitable hydraulic construction machine.
Furthermore, instead of simply taking a weighted mean, we estimate the payload mass by optimizing over an entire movement of the machine, thereby enhancing precision.

Pyrhönen et. al.~\cite{pyrhonen2023mass} addresses this by treating payload estimation as a dual-stage state estimation problem. 
They deduce joint states using only position and pressure data through an Extended Kalman Filter (EKF) that includes hydraulic dynamics, while the payload is determined using inverse dynamics.
The free parameters of a hydraulic model are identified online, minimizing the required calibration ahead of time. 
The absence of inverse dynamics in the kinematics estimator allows the use of an inverse dynamics model in the payload estimation step.
Currently, the presented method is limited to a single joint.
Specific knowledge of the hydraulics setup is required in order to accurately simulate the system behavior.

The idea can similarly be applied to pure rigid body kinematics as shown by Palomba et. al.~\cite{palomba2017kinematic}.
In this approach, estimation is purely informed by kinematic measurements and is modeled through the equations of motion.
Using both, the equations of motion as well as a detailed hydraulics model in a Kalman-filter-based state estimator is shown by Khadim et al.~\cite{khadim_ekf}.
This model enables machine state estimation including hydraulic properties, but limits flexibility and generalizability across machines.
Generally, simulation-based estimation techniques require substantial knowledge about the machine setup and machine-specific parameters, which is not always available in the field.

Although machine learning techniques are sometimes used for weight estimation~\cite{hindman2008dynamic,1206484}, most classical approaches can formulate a direct estimation problem based on joint torques and the equations of motion \cite{Walawalkar_2023,gawlik2017dynamic} with generally good success given the known parameters of the system.
Approaches based on dynamic model identification face the inherent challenge of calibration and parameter estimation~\cite{tafazoli1999identification}.
The coupled nature of a robotic arm, as well as the inability to deconstruct and directly measure the inertial properties, requires proper excitation and model fitting as part of a calibration routine.
The work by Ferlibas and Ghabcheloo~\cite{ferlibas2021load} proposes a lumped model with combined parameters to overcome the issue of not having all parameters individually excitable.
Here, static and dynamic torques are modeled and lumped parameters are fitted to the recorded calibration data.
We extend this approach with a different inertia and joint friction model to support force vector measurements. 
Rather than using the torque difference between two joints to describe the arm dynamics, we formulate a joint-specific model independently applicable to any joint.
Furthermore, we separate the model identification process into the individual contributors, e.g. inertia, gravity, friction to enable easier excitation and a high precision calibration.

Simultaneous estimation with two joints enables \ac{EE} force vector prediction as shown by Palomba~\cite{palomba2019estimation}.
Here, a state estimation approach was used to determine the \ac{EE} forces using known dynamic parameters.
The work presented by Ding~\cite{Ding_Mu_Cheng_Xu_Li_2022} uses the aforementioned identified dynamics approach to also create a reduced order model with identifiable parameters.
It is done on a small robotic manipulator and uses a combined regression matrix for the description of all modeled torques.
Similarly~\cite{kontz2006pressure} shows a pressure based force estimation procedure, but omits the influence of dynamic effects during estimation.
Data-driven approaches such as presented by~\cite{Shen_Wang_Feng_Wang_Fan_2025,Huo_Chen_Wang_2024} show promising results for force estimation in hydraulic systems.
Most of these methods require the presence of ground-truth interaction force measurements for training, which are commonly unavailable on heavy construction machinery.
Learning-based methods inherently rely on data-specific training, making their accuracy and generalization heavily dependent on the diversity and quality of training sets~\cite{Walawalkar_Heep_Frank_Schindler_2020}. Learned methods can be used to improve physics-based methods in their accuracy by compensating for unmodeled effects.

\subsection{Automation of Heavy Machinery}
Automation of heavy construction machinery has a long history in the robotics research community.
Previously automated tasks include excavation~\cite{Egli_Hutter_2022,Terenzi_Hutter_2024,Jin_Ye_Zhang_2023}, trenching~\cite{Jud19AutonomousFreeForm} and grading~\cite{XU2019122} as well as similar tasks in mining and earthworks~\cite{Zhang21AutonomousExcavator,Aoshima_Servin_2024}.
In particular, methods based on model-free \ac{RL} such as~\cite{Egli_Hutter_2022,Lee_Choi_Kim_Moon_Kim_Lee_2022,Spinelli_Egli_Nubert_Nan_Bleumer_Goegler_Brockes_Hofmann_Hutter_2024} achieve good robustness against disturbances and generalize to different environments.
The previously mentioned works all have in common that they directly work on joint level and are specifically tuned for one machine.
A future research direction is the separation of low-level joint control and a machine agnostic \ac{EE} task control only commanding \ac{EE} movement inspired by other research in robotic arms \cite{ha2024umilegs}.
Soil interaction tasks with this architecture will benefit from the measurement of \ac{EE} forces~\cite{DiMaio_1998}.

\section{Model Description and Estimation}
\label{sec:method}
Both, our payload estimation as well as the \ac{EE} force measurement rely on an accurate prediction of the zero-load system torques.
We model the system torques as a combination of gravity, inertia, centrifugal, and frictional effects on the arm elements.
In this section, we present our system torque prediction along the necessary lumping of parameters for calibration.
Furthermore, we show how to compute \ac{EE} force vectors and bucket payloads from the difference between measured and predicted torques.

The naming convention for arm elements, angles, and torques is shown in~\cref{fig:naming}.
\cref{fig:naming} shows the \verb|Boom| joint configuration of a CASE250 excavator with lables for the kinematic joint parameters.
For the sake of conciseness and readability, we assume that the excavator is standing on flat ground.
Cabin pitch deviations can be compensated for by adding the offset to the boom joint angle for gravity and payload torque estimation.
Roll angles require a scaling of the estimated forces and torques, since the force vectors do not remain in the arm plane anymore.

\subsection{Pressures to Torques}
\label{sub:prestotorque}
We compute cylinder forces from pressure readings on the extension $p_1$ and retraction $p_2$ side.
With known cylinder areas of the plunger $A_p$ and rod $A_r$, the resulting force $F$ is described by \cref{eq:cylForce}.
We identify and compensate for mechanical and hydraulic friction in joint space in a later step. 
Therefore, specific friction on the cylinder level is omitted in this stage.
\begin{align}
    F = A_p \cdot p_1 - (A_p - A_r) \cdot p_2
    \label{eq:cylForce}
\end{align}

\begin{figure}
    \centering
    \includegraphics[width=\linewidth]{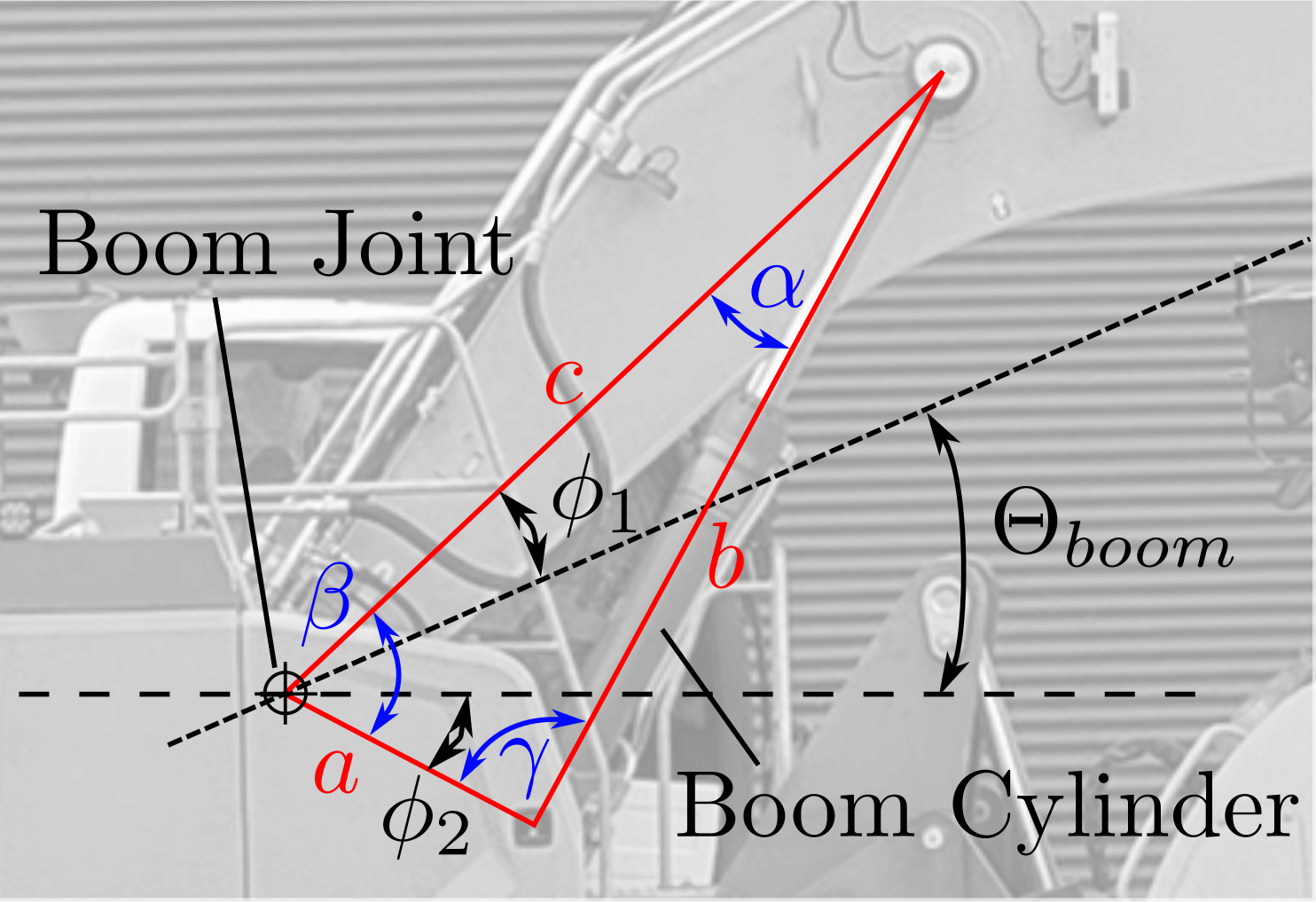}
    \caption{Definition and naming of positions and links in the kinematic setup of the joint.}
    \label{fig:boom_sensitivity}
\end{figure}
The transformation from cylinder force to joint torque can be derived from the kinematic link setup of the joint and is shown here for the basic setup of a boom joint.
With geometric parameters defined as shown in \cref{fig:boom_sensitivity}, joint torque $\tau$ is derived as outlined in \cref{eq:sens1,eq:sens2,eq:sens3}.
All equation parameters are defined in \cref{fig:boom_sensitivity}.
Equivalent equations for the \verb|Stick| joint are derived following the same approach.
\begin{align}
    b &= \sqrt{a^2 + c^2 - 2ac \cdot \cos\left(\Theta_{\text{boom}} + \phi_1 + \phi_2 \right)} \label{eq:sens1}\\
    \eta &= \frac{\partial b}{\partial \Theta_{\text{boom}}} = \frac{a c \sin{\left(\beta + \phi_1 + \phi_2 \right)}}{\sqrt{a^{2} - 2 a c \cos{\left(\beta + \phi_1 + \phi_2 \right)} + c^{2}}} \label{eq:sens2}\\
    \tau &= \eta \cdot F \label{eq:sens3}
\end{align}

\subsection{General model}
We derive the model for zero-load system torques from the equations of motion \cite[Ch.~7]{Siciliano2008-wh}. 
\cref{eq:FullEstim} illustrates the elements that contribute to the measured torque $\tau_\text{m, i}$ in joint $i$. 
When referring to individual joints, the subscript $_i$ identifies the joint \verb|[Boom = 0, Stick = 1, Bucket = 2]| and respective vector index.
We define our base frame origin at the rotational axis of the \verb|Boom| joint, but aligned with gravity.
For conciseness, only the \verb|Boom| joint is described below; the formulation for the \verb|Stick| joint follows a similar but simpler structure.
Since two torque observations suffice for a 2D \ac{EE} force vector estimation, \verb|Bucket| torque measurements are not considered.
The \verb|Bucket| joint is the most imprecise due to the reduced lever length and therefore omitted.
Additionally, during digging and grading motions, it is only used sparsely, which is unfavorable for friction compensation.

\begin{align}
\begin{split}
    \mathbf{\tau}_{\text{m}} = 
    &\underbrace{\mathbf{I}(\mathbf{\Theta}) \bullet \ddot{\mathbf{\Theta}}}_\text{inertia} + 
    \underbrace{\mathbf{\tau}_{\text {fric}}(\dot{\mathbf{\Theta}})}_\text{friction} + 
    \underbrace{\mathbf{\tau_g}(\mathbf{\Theta})}_\text{gravity} \\
    & + \underbrace{\mathbf{h}(\mathbf{\Theta},\dot{\mathbf{\Theta}})}_\text{centripetal} + 
    \underbrace{\mathbf{J(\mathbf{\Theta})}^T\bullet \mathbf{f}_{\text{ext}}}_\text{ext. force}
    \end{split}
    \label{eq:FullEstim}
\end{align}

In~\cref{eq:FullEstim} inertia torques are calculated by multiplication with the rotational acceleration in base frame $\ddot{\mathbf{\Theta}}$.

In literature, friction torques \text{$\mathbf{\tau}_{\text{fric}} \in \mathbb{R}^{3\times1}$} are often velocity-dependent.
We use a friction model that depends on the sign of the velocity as well as the measured joint torque.

\begin{figure}
    \centering
    \includegraphics[width=\linewidth]{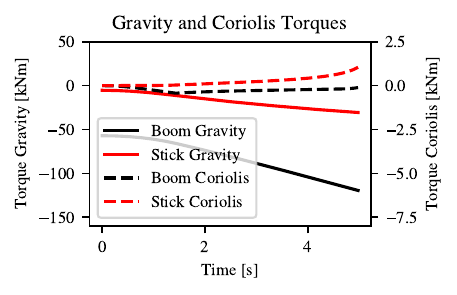}
    \caption{Simulated coriolis and gravity-induced torques for a typical digging motion at \SI{0.7}{\m\per\s} \ac{EE} velocity on M545. }
    \label{fig:grav_corio}
\end{figure}

Gravity-induced torques \text{$\mathbf{\tau_g}(\mathbf{\Theta}) \in \mathbb{R}^{3\times1}$} depend solely on the joint configuration. 
Centripetal and Coriolis forces \text{$\mathbf{h}(\mathbf{\Theta},\dot{\mathbf{\Theta}}) \in \mathbb{R}^{3\times1}$} act on individual links and depend on arm configuration and rotational speeds \text{$\dot{\mathbf{\Theta}} \in \mathbb{R}^{3\times1}$}. 
Due to the need for sustained cabin rotation with varying arm configurations to sufficiently excite centripetal and Coriolis forces during calibration, we opted for an alternative approach. Instead, we approximate the corresponding model using data from our gravity calibration as presented in~\cref{sub:centripetal}.
The coriolis forces resulting from the velocities of the boom, stick and bucket joints are neglected, as their impact is insignificant to achieve the target precision of 1\%. 
We validate this simplification empirically by evaluating Coriolis torques throughout typical digging motions with an \ac{EE} velocity of \SI{0.7}{\m\per\s}, as illustrated in~\cref{fig:grav_corio}. 
Compared to gravity-induced torques, Coriolis and internal centripetal torques \text{$\mathbf{h}(\mathbf{\Theta}, \dot{\mathbf{\Theta}})$} are smaller by more than two orders of magnitude.

Finally, external forces \text{$\mathbf{f}_{\text{ext}} \in \mathbb{R}^{2\times1}$} on the \ac{EE} contribute to joint torques, as described by the configuration-dependent Jacobian \text{$\mathbf{J(\mathbf{\Theta})}\in \mathbb{R}^{2\times3}$}.

\begin{figure}
    \centering
    \includegraphics[width = \linewidth]{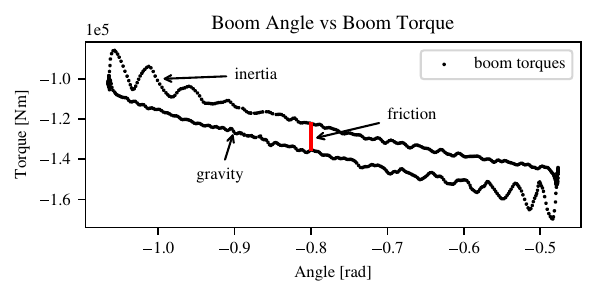}
    \caption{Up - and down motion. Identification of torque contributing effects on CASE250.}
    \label{fig:torqueEffects}
\end{figure}

\cref{fig:torqueEffects} shows the boom torque profile of a simple up and down motion.
The rocking of the machine as well as the cylinder dynamics after starting and stopping induce the ripples at the beginning of the motion and can be described by inertia.
Friction separates the torque for the ascending and lowering path, and gravity torques are the remaining general shape of the profile.

\subsection{Model Parametrization}
Most of the parameters in the torque model from \cref{eq:FullEstim} cannot be separated for identification and therefore must be estimated in lumped form.
We combine the coupled parameters into lumped coefficients $\mathbf{\pi}$.
The following section details the individual components and parameter bundle for gravity ($\mathbf{\pi_g}$), inertia ($\mathbf{\pi_I}$), friction ($\mathbf{\pi_f}$), and centripetal torques ($\mathbf{\pi_c}$).

\subsubsection{Gravity}
\label{sub:gravity}
For a standard excavator with three moving joints along the arm \verb | [Boom, Stick, Bucket] | the decomposition of gravity torques is shown in detail by Ferlibas~\cite{ferlibas2021load} and Tafazoli~\cite{tafazoli1999identification}.
Since the ambiguity of \ac{COG} position and mass cannot be resolved in a quasistatic scenario, two parameters per link are sufficient to describe the torque contribution.
The full expression for gravity torques on the \verb|Boom| joint is listed in~\cref{eq:tau_boom_boom,eq:tau_boom_stick,eq:tau_boom_bucket,eq:tau_total}.
$\tau_{bb}$ is the \verb|Boom| joint torque contribution of the boom link, $\tau_{bs}$ for stick and $\tau_{bbu}$ for the bucket link.
Equivalent, \verb|Stick| gravity torques are described in~\cref{eq:tau_stick_stick,eq:tau_stick_bucket,eq:tau_stick_total} with $\tau_{ss}$ and $\tau_{sbu}$ for stick and bucket link contributions.
The \ac{COG} positions are expressed in the local reference frame with the origin lying on the joint axis and the x-axis pointing towards the succeeding joint.
$m_i$ is the total link mass and $g$ the gravitational acceleration.

\begin{align}
    \begin{split}
        \tau_{bb} = & - g \cdot m_{\text{boom}} \left(cog_{\text{boom} ,x} \cos{\left(\Theta_{\text{boom}} \right)} \right. \\
        & \left. + cog_{\text{boom} ,z} \sin{\left(\Theta_{\text{boom}} \right)}\right)
        \label{eq:tau_boom_boom}
    \end{split} \\
    \begin{split}
        \tau_{bs} = & - g \cdot m_{\text{stick}} \left(cog_{\text{stick} ,x} \cos{\left(\Theta_{\text{boom}} + \Theta_{\text{stick}} \right)} \right. \\
        & + cog_{\text{stick} ,z} \sin{\left(\Theta_{\text{boom}} + \Theta_{\text{stick}} \right)} \\
        & \left. + l_{\text{boom}} \cos{\left(\Theta_{\text{boom}} \right)}\right)
        \label{eq:tau_boom_stick}
    \end{split} \\
    \begin{split}
        \tau_{bbu} = & - g \cdot m_{\text{bucket}} \left(cog_{\text{bucket} ,x} \right. \\
        & \left. \cos{\left(\Theta_{\text{boom}} + \Theta_{\text{bucket}} + \Theta_{\text{stick}} \right)} \right. \\
        & + cog_{\text{bucket} ,z} \sin{\left(\Theta_{\text{boom}} + \Theta_{\text{bucket}} + \Theta_{\text{stick}} \right)} \\
        & + l_{\text{boom}} \cos{\left(\Theta_{\text{boom}} \right)} \\
        & \left. + l_{\text{stick}} \cos{\left(\Theta_{\text{boom}} + \Theta_{\text{stick}} \right)} \right)
        \label{eq:tau_boom_bucket}
    \end{split} \\
    \tau&_{\text{grav, boom}} = \tau_{bb} + \tau_{bs} + \tau_{bbu}
    \label{eq:tau_total}
\end{align}

\begin{align}
    \begin{split}
        \tau_{ss} = & - g \cdot m_{\text{stick}} \left(cog_{\text{stick} ,x} \cos{\left({\Theta_{\text{boom}}+\Theta_\text{stick}} \right)} \right. \\
        & \left. + cog_{\text{stick} ,z} \sin{\left(\Theta_{\text{boom}} + \Theta_{\text{stick}} \right)}\right)
    \end{split}
    \label{eq:tau_stick_stick} \\
    \begin{split}
        \tau_{sbu} = & - g \cdot m_{\text{bucket}} \left( cog_{\text{bucket},x}  \right. \\
        & \left. \cos\left({\Theta_{\text{boom}} + \Theta_\text{stick}} + \Theta_{\text{bucket}} \right) \right. \\
        & + cog_{\text{bucket},z} \sin\left(\Theta_{\text{boom}} + \Theta_\text{stick} + \Theta_{\text{bucket}} \right) \\
        & + l_{\text{stick}} \cos\left({\Theta_{\text{boom}} + \Theta_\text{stick}} \right) \left. \right)
    \end{split}
    \label{eq:tau_stick_bucket} \\
    \tau&_{\text{grav, stick}} = \tau_{ss} + \tau_{sbu}
    \label{eq:tau_stick_total}
\end{align}

\cref{eq:lumped_grav,eq:lumped_grav_stick} show the lumped forms that combine \ac{COG}, mass, subsequent joint mass and link length for \verb|Boom| and \verb|Stick| joints.
\begin{align}
    \begin{split}
        \tau_{\text{grav, boom}} = &  \pi_{bg, 1} \cdot \cos{(\Theta_{\text{boom}})} + \\
                                    & \pi_{bg, 2} \cdot \sin{(\Theta_{\text{boom}})}+ \\
                                    & \pi_{bg, 3} \cdot \cos{(\Theta_{\text{boom}} + \Theta_{\text{stick}})}+ \\
                                    & \pi_{bg, 4} \cdot \sin{(\Theta_{\text{boom}} + \Theta_{\text{stick}})}+ \\
                                    & \pi_{bg, 5} \cdot \cos{(\Theta_{\text{boom}} + \Theta_{\text{stick}} + \Theta_{\text{bucket}})}+ \\
                                    & \pi_{bg, 6} \cdot \sin{(\Theta_{\text{boom}} + \Theta_{\text{stick}} + \Theta_{\text{bucket}})}
    \end{split}
    \label{eq:lumped_grav}
\end{align}
\begin{align}
    \begin{split}
        \tau_{\text{grav, stick}} = &  \pi_{sg, 7} \cdot \cos{(\Theta_{\text{boom}} + \Theta_{\text{stick}})} + \\
                                    & \pi_{sg, 8} \cdot \sin{(\Theta_{\text{boom}} + \Theta_{\text{stick}})} + \\
                                    & \pi_{sg, 9} \cdot \cos{(\Theta_{\text{boom}} + \Theta_{\text{stick}} + \Theta_{\text{bucket}})} + \\
                                    & \pi_{sg,10} \cdot \sin{(\Theta_{\text{boom}} + \Theta_{\text{stick}} + \Theta_{\text{bucket}})}
    \end{split}
    \label{eq:lumped_grav_stick}
\end{align}

\subsubsection{Inertia}
\label{sub:Inertia}
The arm inertia with respect to the observed joint has constant and joint position-dependent contributors.
The constant part combines the rotational inertia of each link around its own \ac{COG} as shown for the \verb|Boom| joint in \cref{eq:inertia_arm}.
Following Steiner's Theorem, arm inertia is further dependent on the link masses and distances between the axis of rotation and each link's \ac{COG}.

Since $d_{bo}$ (distance from center of rotation to boom \ac{COG}) and $m_{boom}$ as well as all highlighted coefficients in distance~\cref{eq:distance_stick,eq:distance_bucket} do not depend on the joint positions, they also become part of the constant parameter.
This only leaves $m_{\text{stick}} \cdot d_{\text{st}}^2 + m_{\text{bucket}} \cdot d_{\text{bu}}^2$ to be parameterized. 
The lumped inertia model for identification regarding the \verb|Boom| joint is shown in~\cref{eq:inertiaModel}.
\cref{eq:inertiaModelStick} shows the lumped model regarding the \verb|Stick| joint.
\begin{align}
    \begin{split}
            I_b = & \underbrace{m_{\text{boom}} \cdot d_{\text{bo}}^2 + m_{\text{stick}} \cdot d_{\text{st}}^2 + m_{\text{bucket}} \cdot d_{\text{bu}}^2}_{\text{Steiner's Theorem rotation around offset}} + \\
            &\underbrace{I_{zz, \text{boom}} + I_{zz, \text{stick}} + I_{zz, \text{bucket}}}_{\text{Rotational inertia part of each link}}                
    \end{split}
    \label{eq:inertia_arm}
\end{align}

\begin{align}
    \begin{split}
    d_{st}^2 = &\eqHigh{l_{\text{boom}}^{2}} + \eqHigh{cog_{\text{stick} ,x}^{2}} + \eqHigh{cog_{\text{stick} ,z}^{2}} + \\
                &  2 cog_{\text{stick} ,x} \cdot l_{\text{boom}} \cos{\left(\Theta_{\text{stick}} \right)} + \\
                & 2 cog_{\text{stick} ,z} \cdot l_{\text{boom}} \sin{\left(\Theta_{\text{stick}} \right)}                   
    \end{split}
    \label{eq:distance_stick}
\end{align}
\begin{align}
    \begin{split}
    d_{bu}^2 = & \eqHigh{cog_{\text{bucket} ,x}^{2}} + \eqHigh{cog_{\text{bucket} ,z}^{2}} + \\ 
               & 2 cog_{\text{bucket} ,x} \cdot l_{\text{boom}} \cdot \cos{\left(\Theta_{\text{bucket}} + \Theta_{\text{stick}} \right)} + \\
               & 2 cog_{\text{bucket} ,x} \cdot l_{\text{stick}} \cdot \cos{\left(\Theta_{\text{bucket}} \right)} + \\
               & 2 cog_{\text{bucket} ,z} \cdot l_{\text{boom}} \cdot \sin{\left(\Theta_{\text{bucket}} + \Theta_{\text{stick}} \right)} + \\
               & 2 cog_{\text{bucket} ,z} \cdot l_{\text{stick}} \cdot \sin{\left(\Theta_{\text{bucket}} \right)} + \\
               & \eqHigh{l_{\text{boom}}^{2}} + 2 l_{\text{boom}} l_{\text{stick}} \cos{\left(\Theta_{\text{stick}} \right)} + \eqHigh{l_{\text{stick}}^{2}}
    \end{split}
    \label{eq:distance_bucket}
\end{align}

\begin{align}
    \label{eq:inertiaModel}
    \begin{split}
        I_{b} =& \pi_{bI, 1} + \pi_{bI, 2} \cos{(\Theta_{\text{stick}})} + \pi_{bI, 3} \sin{(\Theta_{\text{stick}})} + \\
                                    & \pi_{bI, 4} (2 l_{\text{boom}} \cos{(\Theta_{\text{stick}} + \Theta_{\text{bucket}})} + \\
                                    & 2 l_{\text{stick}} \cos{(\Theta_{\text{bucket}})}) + \\
                                    & \pi_{bI, 5} (2 l_{\text{boom}} \sin{(\Theta_{\text{stick}} + \Theta_{\text{bucket}})} + \\
                                    & 2 l_{\text{stick}} \sin{(\Theta_{\text{bucket}})})
    \end{split}
\end{align}

\begin{align}
    \label{eq:inertiaModelStick}
    \begin{split}
        I_{s} =& \pi_{sI, 1} + \pi_{sI, 2} \cos{(\Theta_{\text{bucket}})} + \pi_{sI, 3} \sin{(\Theta_{\text{bucket}})}
    \end{split}
\end{align}

\subsubsection{Friction}
\label{sub:friction}
We observed that friction in joint space has little to no dependence on the magnitude of the joint velocity.
Similarly to Bonchis~\cite{bonchis1999pressure} we use a friction model that is dependent on torque and direction of motion.
The dependency on joint torque scales with increasing pressure on gaskets, displacement of lubrication, and increase in hydraulic friction from pressure.
As chamber pressures are measured right at the cylinder, no significant contribution of flow-dependent components is expected.
A different set of parameters is used for lifting and lowering motions, making the model asymmetric.
Describing the friction torque $\tau_{f, i}$ as a linear function of the measured torque $\tau_{m, i}$ as in \cref{eq:fricTorque} has proved sufficient.
\begin{align}
    \tau_{f, i} =& \pi_{f, i, 1} \cdot \tau_{m, i} + \pi_{f, i, 2}
    \label{eq:fricTorque}
\end{align}

\subsubsection{Centripetal}
\label{sub:centripetal}
All preceding considerations were constrained to in-plane motions of the arm joints.
While during digging, the cabin is static and thus no centripetal forces are excited, a typical loading cycle includes motion from the cabin joint.
Therefore, the payload estimation feature greatly benefits from centripetal torque compensation contrary to the \ac{EE} force estimation.
Estimating the shovel payload in a nominal digging cycle requires consideration of centripetal forces during slewing motion due to high turning rates.
Exciting the slew $\dot{\Theta}_{\text{cab}}$ continuously and long enough for different arm configurations as necessary for calibration is a challenging task in reality.

The equations for centripetal torques $\mathbf{\tau_{c, \text{boom}}}$ largely overlap with the already parameterized gravity torque equations.
Therefore, we can approximate centripetal torques with the previously identified parameters from \cref{sub:gravity} and one additional scaling parameter $\pi_c$. Re-ordering of the identified gravity parameters predicts induced torques from lateral acceleration, as it is the case from centripetal forces.
$\pi_{c}$ compensates for the error introduced by omitting the distance to secondary links in the kinematic chain.
Centrifugal torques are calculated by the model shown in \cref{eq:centtorque}.
The free parameter $\pi_c$ can be identified on a single dataset which includes rotation of the cabin.
\begin{align}
    \begin{split}
        \tau_{c, \text{boom}} =& \dot{\Theta}_{\text{cab}}^2 \cdot \pi_c \cdot \\
                & (\pi_{bg, 2} \cdot \cos{(\Theta_{\text{boom}})} + \\
                & \pi_{bg, 1} \cdot \sin{(\Theta_{\text{boom}})}+ \\
                & \pi_{bg, 4} \cdot \cos{(\Theta_{\text{boom}} + \Theta_{\text{stick}})}+ \\
                & \pi_{bg, 3} \cdot \sin{(\Theta_{\text{boom}} + \Theta_{\text{stick}})}+ \\
                & \pi_{bg, 6} \cdot \cos{(\Theta_{\text{boom}} + \Theta_{\text{stick}} + \Theta_{\text{bucket}})}+ \\
                & \pi_{bg, 5} \cdot \sin{(\Theta_{\text{boom}} + \Theta_{\text{stick}} + \Theta_{\text{bucket}})})
    \end{split}
    \label{eq:centtorque}
\end{align}

\subsection{Force and Payload Measuring}
Both our payload- and force vector estimators are comparing predicted zero-load system torques from the previously described model to measured torques as described in~\cref{sub:prestotorque}.
The online force vector estimation uses \verb|Boom| and \verb|Stick| torques to calculate the magnitude and direction of the contact forces on the shovel blade. 
This is done for every time step and is updated live. 
Therefore, no optimization is used.

The payload estimation utilizes only \verb|Boom| torques, and considers measurements recorded during a lifting or lowering motion of the excavators' arm.
The payload is determined as the solution to an optimization problem, taking the whole trajectory as input by matching the measured torques with the presented model.
Since bucket payload remains constant, optimizing for the payload over a whole motion rather than just one sample greatly increases precision.
A single payload value is computed for the entire motion from start to finish. 

\subsubsection{Force Vector Estimation}
For the force vector estimation, we are assuming an empty shovel and thus no inertia or gravity contribution of a potential payload.
Subtracting the predicted zero load torque from the measurement yields $\Delta \tau_{\text{boom}}$ and $\Delta \tau_{\text{stick}}$.
The Jacobian \text{$\mathbf{J} \in \mathbb{R}^{2\times2}$} used to retrieve the in-plane \ac{EE} forces is computed from the partial derivative of the kinematic translation \text{$\mathbf{r} \in \mathbb{R}^{2\times1}$} to the blade with respect to the boom and stick joint. The force calculation from torques in \cref{eq:extForces} is used.
\text{$\mathbf{T}_{\text{boom}}^{\text{blade}} \in \mathbb{R}^{3 \times 3}$} is the homogeneous coordinate transform from the boom joint to the blade.
The translation \text{$\mathbf{r} \in \mathbb{R}^{2\times1}$} is a function of all joint angles \text{$\mathbf{\Theta} \in \mathbb{R}^{3\times1}$}.
\begin{align}
    \mathbf{T_{\text{boom}}^{\text{blade}}(\mathbf{\Theta})} =& \begin{bmatrix}\mathbf{C(\mathbf{\Theta})}& \mathbf{r(\mathbf{\Theta})} \\ \mathbf{0} & 1\end{bmatrix}
\end{align}
    
\begin{align}
    \label{eq:extForces}
    \mathbf{f_{ext}} =& - \mathbf{J}^{-1, T}(\mathbf{\Theta}) \bullet \left[ \Delta \tau_{\text{boom}}, \Delta \tau_{\text{stick}} \right]^T
\end{align}

\subsubsection{Payload Estimation}
Motion of the boom joint is required to overcome stiction as well as to ensure that the estimated friction torque is valid.
Due to link kinematics and payload lever length, the measurement sensitivity changes within the workspace.
Over the normal working envelope, the pressure to payload sensitivity of our experimental machines changes approximately by a factor of 3.
Joint positions, velocities, accelerations, and torques are recorded during motion.
A batch optimization problem~\eqref{eq:weightMinFunc} to minimize $\Delta \tau_{\text{boom}} - \tau_{\text{pl, boom}}$ with the only free parameter being the payload mass $m_{\text{pl}}$ is being solved using the Nelder Mead algorithm. 
Since batch optimization only has to be executed once per motion cycle, its runtime is not critical and only has to be finished by the next episode start.
Depending on the operator and task, recorded motion episodes contain only a couple hundred samples, with Nelder Mead typically converging after only a few iterations, staying at execution times significantly below \SI{1}{\s} on embedded hardware.

$\tau_{\text{pl, boom}}$ includes the gravity torque of a mass located at the shovel center $\mathbf{\tau}_{\text{pl, boom, g}}$, as well as added inertia torques at the \verb|Boom| joint $\tau_{\text{pl, i}}$ and slew centrifugal torques of a point mass at the same location $\mathbf{\tau}_{\text{pl, boom, c}}$, as shown in \cref{eq:pqyloadT}.

\begin{align}
\label{eq:weightMinFunc}
    \min_{m_{\text{pl}}} \Delta &\tau_{\text{boom}} - \tau_{\text{pl, boom}}(\mathbf{\Theta}, \dot{\mathbf{\Theta}}, \ddot{\mathbf{\Theta}}, m_{\text{pl}})\\
    \mathbf{\tau}_{\text{pl, c}} =& \mathbf{J}^T \bullet (r_x(\mathbf{\Theta}) \cdot m_{\text{pl}} \cdot \dot{\Theta}_{cab}^2 \cdot[1, 0]^T)\\
    \mathbf{\tau}_{\text{pl, g}} =& -\mathbf{J}^T \bullet ( m_{\text{pl}}\cdot[0, -g]^T)\\
    \tau_{\text{pl, boom, i}} =& \ddot{\Theta}_{\text{boom}} \cdot |\mathbf{r}(\mathbf{\Theta})|^2 \cdot m_{\text{pl}}\\
    \tau_{\text{pl, boom}} = &\tau_{\text{pl, boom, c}} + \tau_{\text{pl, boom, g}} +\tau_{\text{pl, boom, i}}
    \label{eq:pqyloadT}
\end{align}

\subsection{Sensitivity analysis}
\begin{figure}
    \centering
    \includegraphics[width =.8\linewidth]{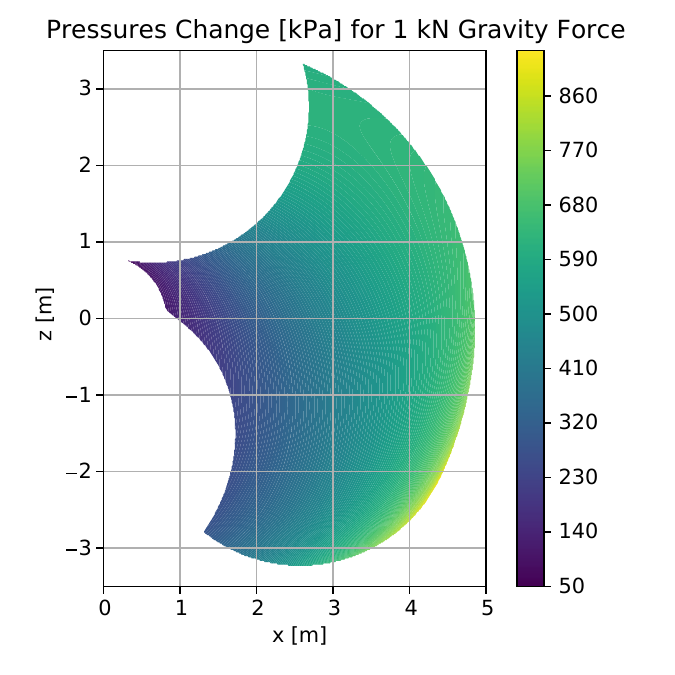}
    \caption{Boom chamber pressure change for \SI{1}{\kN} \ac{EE} gravity force over the workspace of the M545 machine. Color shows kinematic sensitivity. The prismatic joint of the M545 remains locked for our experiments.}
    \label{fig:pressureSensitivity}
\end{figure}

 \cref{fig:pressureSensitivity} illustrates the workspace pressure sensitivity of the \verb|Boom| joint in response to a $\sim$~\SI{100}{\kg} payload change on a Menzi Muck M545 excavator. 
A significant variation in sensitivity of approximately a factor of 3 is observed in the normal working envelope of the machine.
This analysis in comparison with the other joints identifies the \verb|Boom| joint as the most suitable for payload estimation and helps to select a pressure sensor accuracy for the target full-scale accuracy of the system.

\section{Calibration}
The presented identification of free parameters is designed to retrofit existing machines without requiring costly disassembly.
All plots in this section display data from a Menzi Muck M545 excavator~\cite{Jud_Kerscher_Wermelinger_Jelavic_Egli_Leemann_Hottiger_Hutter_2021}.

The only prerequisite is knowledge about the kinematic link lengths as well as the joint sensitivity factor. 
The joint sensitivity factor (cylinder force to joint torque) can be inferred from the joint kinematics.
Calibration, analogous to operation, requires joint angles, velocities, accelerations, and pressure readings on the cylinders. 

If unknown, calibration begins with identifying the size of the cylinder plunger, which is typically inaccessible to the user. 
While the diameter of the rod can be measured geometrically, the size of the plunger must be identified from data.

\begin{figure}
    \centering
    \includegraphics[width = \linewidth]{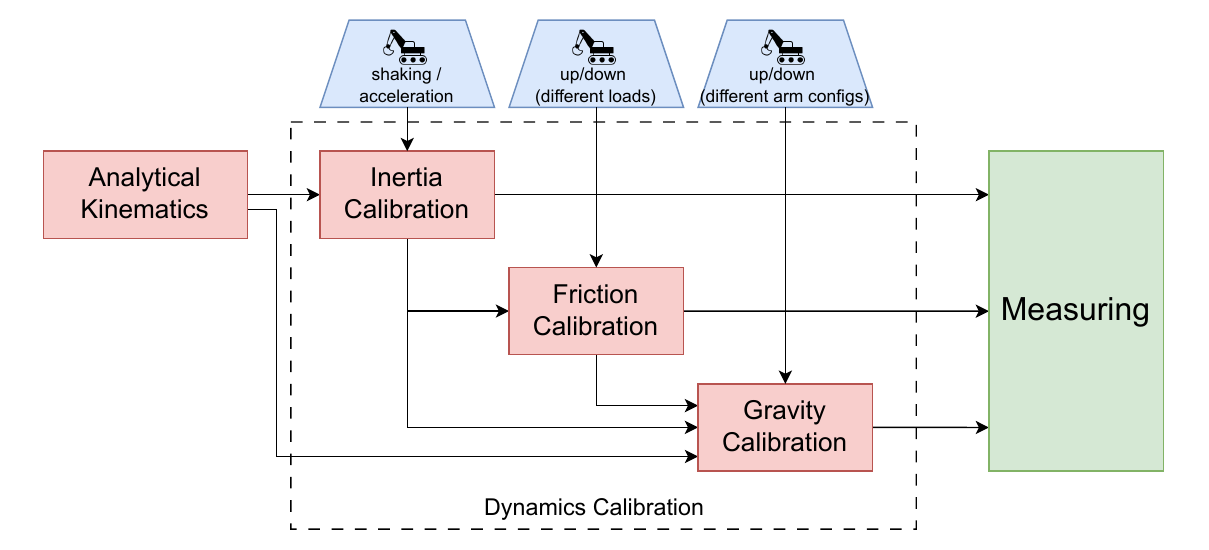}
    \caption{Flow diagram of the calibration sequence.}
    \label{fig:calibPipeline}
\end{figure}

With joint torque measurements, the next steps involve identifying the model for static and dynamic torques of the unloaded arm. 
This process is done individually for each joint with individual calibration steps shown in \cref{fig:calibPipeline}. 
Although unified calibration of all joints for all dynamic and static parameters $\mathbf{\pi}$ is certainly possible, we achieve better results with a staged calibration process.
Each stage uses a dataset that specifically excites the identifiable parameters.
The parameter calibration routine must be performed in the presented order as later steps build upon previously acquired parameters. 
Each of the three steps requires different arm movements, necessitating three individual datasets per joint for a complete calibration. 
The entire process takes no longer than \SI{30}{\min} of machine time.

\subsection{Cylinder Plunger Size}
\begin{figure}
    \centering
    \includegraphics[width=\linewidth]{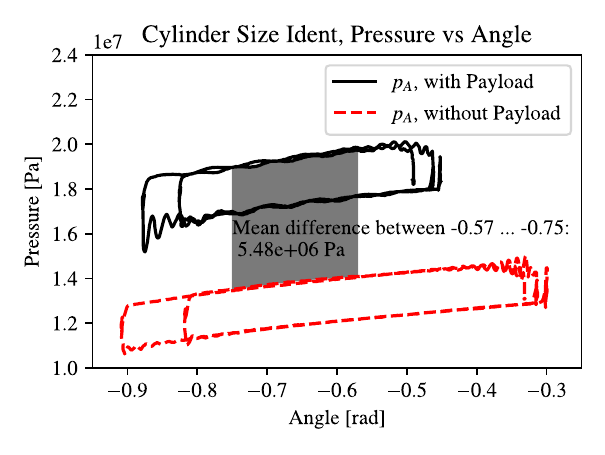}
    \caption{Cylinder size identification datasets. Only a steady, overlapping path of the motions are used for identification. Data from M545.}
    \label{fig:cylinderIdent}
\end{figure}

Since the diameters of the plunger and rod are needed for all other calibrations, no information regarding link masses, \ac{COG} or moments of inertia is available at this point.
This is overcome by recording two raising and lowering joint motions, with all other joints remaining in the exact same configuration.
The only difference between the two datasets is the addition of a known payload suspended on the shovel blade.
The datasets should include multiple smooth up and down motions.
\cref{fig:cylinderIdent} shows the comparison of the two datasets for the plunger side of the boom cylinder on a Menzi Muck M545.
By exclusively considering the difference between the two datasets, all torques originating from the arm itself are compensated for as long as the motion is steady.

The added payload creates a predictable torque in the joint, which can be resolved as shown in \cref{eq:cylIdent}. 
Here, $p_2^{wo}$ denotes the pressure on the loaded side \textit{without} the payload, and $p_2^{w}$ the corresponding pressure \textit{with} the additional payload at a given point in time.
$\Theta_i$ is the angle of the joint, $\Theta_m$ the angle offset to the payload, and $|r|$ the distance between the joint and the payload.
\begin{align}
    A_p = \frac{A_r \cdot( \eta (\Theta_i) (p_2^{wo} - p_2^{w} )) + g |r| m_{\text{pl}} \cos(\Theta_i - \Theta_m)}{\eta(\Theta_i)((p_1^{w} - p_2^{w})-(p_1^{wo} - p_2^{wo}))}
    \label{eq:cylIdent}
\end{align}
We select a overlapping region of the quasi static motion for calibration.
The average estimated plunger area is subsequently used for querying a database of known cylinder sizes from different norms and industry standards to identify the correct cylinder type.

\subsection{Inertia}
Our arm parameter estimation begins with inertia.
At the beginning and end of each motion, the machine shakes significantly, exciting the inertia torques. 
Since typical machine rocking eigenfrequencies are very close to the acceleration signal during normal operation, they cannot be sufficiently low-pass filtered.
Starting with identification of inertia terms simplifies subsequent friction and gravity identification, as the oscillations span a significant portion of the motion, as shown in \cref{fig:torqueEffects}.

The lack of knowledge about gravity and friction forces prohibits direct identification of the free parameters $\mathbf{\pi_I}$ in the inertia model $I_b$ as introduced in~\cref{sub:Inertia}.
Since excitation of friction and gravity occurs quasistatically, we can separate the inertia-induced signal in the frequency domain rather than in the time domain. 
Minimizing the accumulated \ac{PSD} from \SI{0.5}{\Hz} to \SI{3}{\Hz}, as described in~\cref{eq:optimProblemInertia}, yields a set of $\mathbf{\pi_I}$ parameters that describe the inertia torques in the identified joint.
$\mathscr{F}$ denotes the Fourier transformation.
The optimization band was empirically determined by observing accelerations and shaking of different excavators in multiple weight classes.

\begin{align}
    \begin{split}
        \min_{\mathbf{\pi_I}} &\int_{\SI{0.5}{\Hz}}^{\SI{3}{\Hz}}S_{t}(f)\\
        \text{with }& S_t(f) : =  \mathscr{F}(\tau_{m}(t) - (I_b(\mathbf{\pi_i}, t) \cdot \ddot{\Theta}_i))
    \end{split}
    \label{eq:optimProblemInertia}
\end{align}

\begin{figure}
    \centering
    \includegraphics[width = \linewidth]{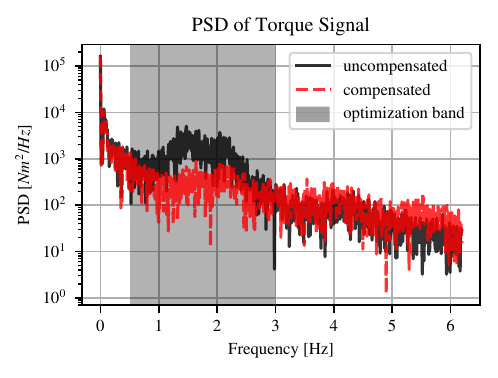}
    \caption{Comparison of the \ac{PSD} of raw measurements with inertia compensated measurements.}
    \label{fig:inertiaPSD}
\end{figure}

\cref{fig:inertiaPSD} shows the comparison of the torque signals \ac{PSD} before and after subtraction of estimated Inertia torques in frequency domain.

Calibration requires a dataset with high and repeated joint accelerations.
To excite all $\mathbf{\pi_I}$ parameters, the arm configuration must change after each motion to cover the entire workspace of the machine.
Using the natural rocking of the excavator after hard accelerations of one arm joint is the easiest way to excite inertia-induced torques. 
A dataset with approximately 10 full up-and-down motions, involving a rocking machine and different arm configurations, has proved sufficient for reliable identification.

\subsection{Friction}
This methods friction identification uses the previously identified inertia model to remove oscillations from machine rocking and joint acceleration torques from the dataset prior to identification. 
The friction model depends only on the measured joint torque and the direction of motion. 
Different torques are excited by changing the arm configuration, thereby altering the gravity loads on the joint. 
To compensate for noise and any remaining inertia residuals, multiple joint movements with the same arm configuration are repeated before changing configurations. 
Four different configurations, ranging from completely curled in to fully extended, have proven to be sufficient for identification.

\begin{figure}
    \centering
    \includegraphics[width = \linewidth]{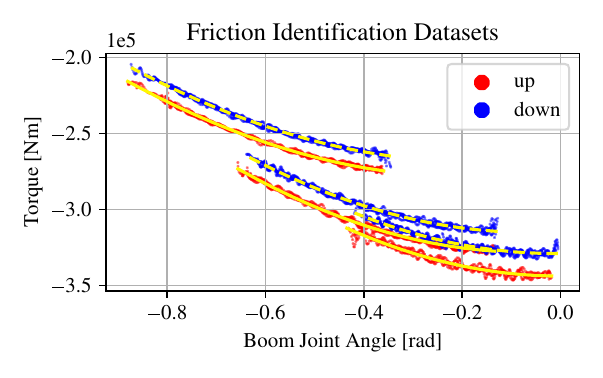}
    \caption{Torque trajectories for friction identification with the second-order fit overlaid.}
    \label{fig:gravityTorqueTrajectories}
\end{figure}

\begin{figure}
    \centering
    \includegraphics[width=\linewidth]{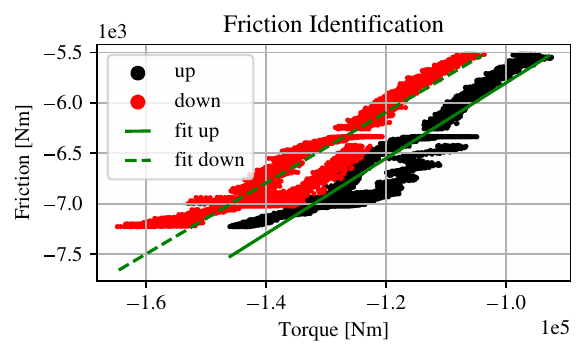}
    \caption{Torque resulting from friction plotted against the measured torque from the Joint. Measured torques include friction, gravity, and dynamic effects.}
    \label{fig:frictionIdent}
\end{figure}

\cref{fig:gravityTorqueTrajectories} shows the torque trajectories for lifting and lowering motions of three different arm configurations. The datasets include motions of different velocities, yet no difference in the torque profile is visible. A second-order function is fitted to the torque data of each set's rising and lowering motions to average over the individual motions. The fits are represented as dashed (lowering) and solid (rising) lines in the plot. The distance between associated lines is caused by friction.
Observing the torque difference between lifting and lowering as a function of the measured torque reveals an approximately linear dependence. This implies that the friction in each joint varies with how much load rests on the joint itself. 
\cref{fig:frictionIdent} shows the described linear dependency.
The identified first order function parameters are used to predict the friction torques in the final model

\subsection{Gravity}
\begin{figure}
    \centering
    \includegraphics[width=\linewidth]{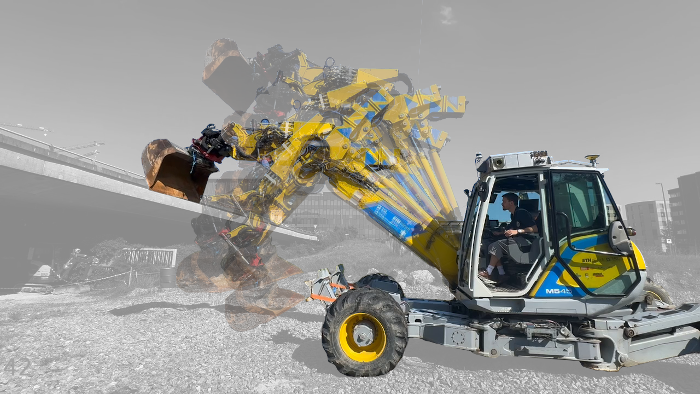}
    \caption{Gravity Calibration movement of M545}
    \label{fig:gravCalibM545}
\end{figure}

Calibrating for the free parameters in the gravity model $\pi_g$ directly works by optimizing the residual between predicted and measured torques.
The complete model includes inertia, friction, and gravity torques.
A dataset for calibration should again include multiple lifting and lowering cycles with a changing arm configuration.
This dataset should not include excessive shaking contrary to the inertia dataset, and more arm configurations spanning the whole workspace of the machine are needed compared with the friction calibration.
\cref{fig:gravCalibM545} shows the range of motion for a gravity calibration on the Menzi Muck M545.

The optimization problem formulated in \cref{eq:gravityIdent} using the gravity model previously layed out in \cref{sub:gravity} can be solved by least-squares optimization.
\begin{align}
    \min_{\mathbf{\pi_g}} \tau_m - (I_i(\mathbf{\Theta}) \ddot{\Theta}_{i} + \tau_{\text {fric}, i} + \tau_{g,i}(\mathbf{\Theta, \pi_g}))
    \label{eq:gravityIdent}
\end{align}

\section{Experiments}
The presented method is deployed and evaluated on two different excavators of different weight classes to ensure that the technique is machine-agnostic.
A 12t Menzi Muck M545 walking excavator as well as a 25t CASE250 are used for evaluation.
For the sake of comparability, the prismatic joint of the M545 is locked during experiments.
Platform wise, the M545 is much smaller, lighter and more agile than the 25t standard excavator.
Higher joint accelerations are possible, while significant oscillations originating from air-filled tires pose a particular challenge on this platform.
Table \ref{tab:excavator_comparison} compares the main metrics of the machinery used.

\begin{table}[ht]
\centering
\caption{Comparison of Menzi Muck M545 and CASE 250 Excavators}
\label{tab:excavator_comparison}
\begin{tabular}{|l|c|c|}
\hline
\textbf{Parameter} & \textbf{M545} & \textbf{CASE 250} \\
\hline
Weight                & 13.5\,t          & 25\,t        \\
Engine Power          & 115\,kW          & 138\,kW      \\
Overall Length        & 6.5\,m           & 5.3\,m       \\
Overall Width         & 2.3\,m           & 3.2\,m       \\
Boom Length           & 2.8\,m           & 5.8\,m       \\
Stick Length          & 2.1\,m           & 3.0\,m       \\
Bucket Length         & 1.4\,m           & 1.6\,m       \\
Bucket Capacity       & 0.54\,m$^3$      & 0.8\,m$^3$   \\
Maximum Reach         & 8.7\,m\footnotemark[1] & 10.8\,m \\
\hline
\end{tabular}
\footnotetext[1]{Reach measured with telescoping stick}
\end{table}

The results of the CASE250 evaluation are shown in this paper in favor of conciseness.
Results from the M545 are matching in relative error and quality.
The major difference between the platforms are the air-filled tires on M545.
Oscillating dynamics can be exploited for inertia excitation, while at the same time, a correctly identified inertia is significantly more important during operation.
Even normal digging cycles excite strong accelerations which have to be compensated for an accurate \ac{EE} force reading.
Motions on CASE250 are slower and smoother in comparison, which ultimately did not lead to a different relative accuracy.

First, we evaluate the accuracy of the live zero-load torque prediction as it serves as a foundation for payload and force measurements.
Furthermore, the accuracy of the payload estimation algorithm is verified and compared against a quasistatic reference method as well as a commercial solution.
Finally, we validate the force vector measurements by observing the force from a load suspended from the shovel.

\subsection{Experiment Setup}
Both machines used for testing are equipped with a Leica iCON kinematic sensing system~\cite{ICON_website}.
It provides us with \SI{50}{\Hz} joint position and velocity readings.
Both are derived internally from inertial sensors (Leica MSS40x) (Accelerometer, Gyroscope) mounted on the individual links of the arm.
We compute joint accelerations by fitting a first-order function to the last five measurements.
In order to re-align the phase of the other measurements with the delayed accelerations, we average the last five measurements of joint position, velocity, and pressures accordingly.

For cylinder pressure readings, we equipped the boom and stick cylinders with Wika P30 pressure sensors~\cite{WIKA_website}.
No hydraulic elements remain between the cylinder chambers and pressure sensors to avoid the inclusion of un-modeled hydraulic components.

\subsection{Basic Signal Comparison}
\begin{figure}
    \centering
    \includegraphics[width = \linewidth]{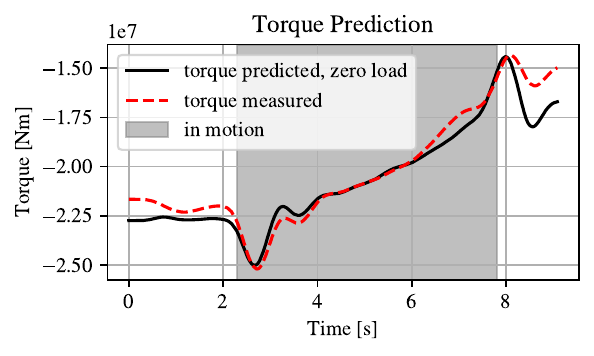}
    \caption{Examplary zero load torque prediction on boom joint of M545}
    \label{fig:zeroLoadTorquePrediction}
\end{figure}

The first experiment demonstrates a simple \SI{5}{\s} lifting trajectory with high accelerations at the start and end. 
\cref{fig:zeroLoadTorquePrediction} shows the measured and predicted torque profiles for this motion. 
The non-highlighted areas indicate periods without joint motion, thus not accurately predicting the unmodeled static friction effects within the joint. 
Once the joint is in motion, the measured torque is accurately tracked by the estimator, even during high acceleration phases.

\subsection{Gravity Torque Workspace}
\begin{figure}
    \centering
    \includegraphics[width = .8\linewidth]{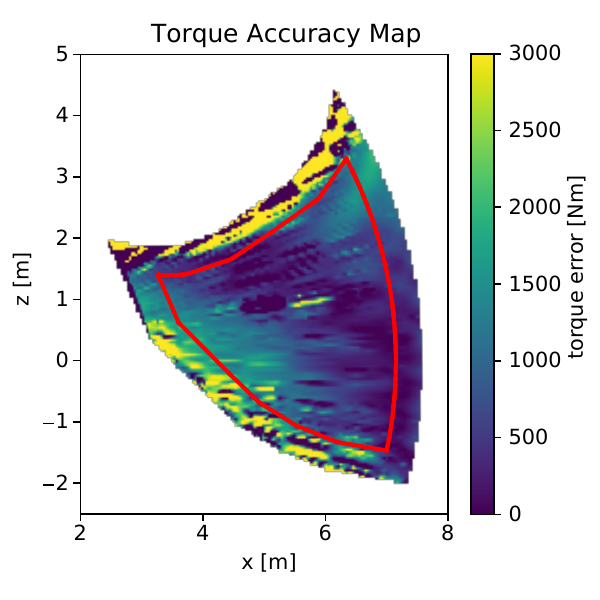}
    \caption{Workspace analysis of torque prediction accuracy on CASE250. Data interpolated from 10 consecutive lifting motions spread across the whole workspace of the machine.}
    \label{fig:torquePredictionAccuracyMap}
\end{figure}

To ensure consistently accurate system torque prediction across the machine's entire workspace, a dataset of 15 trajectories covering the accessible part of the workspace was collected. For each point, the torque prediction error in the \verb|Boom| joint was calculated. \cref{fig:torquePredictionAccuracyMap} shows an interpolated version of this error map across the accessible workspace.

Torque offsets from static friction and machine oscillation cause an inflated error at the start and end of each motion, visible as brighter areas at the bottom and top. The top area also shows artifacts from interpolation in sparse regions. To measure the average error, a reduced workspace was used to exclude the static and high-acceleration parts.
The average error in the prediction of gravity torque in the CASE250 machine is \SI{644}{\newton\meter}, with a 98th percentile of \SI{2186}{\newton\meter}. This corresponds to a maximum instantaneous weight error of \SI{36}{\kg} at a nominal working point of \SI{6}{\m} shovel distance.

\subsection{Payload Estimation}
Since the weight estimation pipeline compares measured torques with predicted zero-load torques, the magnitude of the estimation error is independent of the payload. 
To quantify the error of the offline weight estimation pipeline, we collected weighing trajectories spanning the entire machine workspace. 
The datasets include maximum power accelerations, multi-joint motions, fast slewing motions, and very short trajectories of \SI{0.5}{\s} duration to benchmark the system's lower bound performance in realistic deployment scenarios.
We report the measurement standard deviation, calculated from all experiments along with the mean value or error to aid in interpretability of the systems' consistency.

As a comparison, we implemented the quasistatic hypersurface-based estimator described by Bennett~\cite{bennet2014payload}. 
Additionally, we benchmarked our system against a commercially available payload estimation system under similar conditions on a 25t machine. 
The results are shown in \cref{tab:comparison}.
Both, the hypersurface-based and commercial estimator run online with an averaged output value at the end of the motion. 
As the real payload does not significantly change during nominal in-air motions, we compare the averaged online payload to our post-motion optimized estimate.

\begin{figure}
    \centering
    \includegraphics[width=\linewidth]{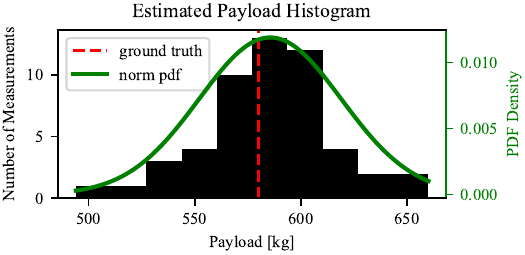}
    \vspace*{-5pt} 
    \caption{Histogram of 52 weight measurements on CASE250}
    \label{fig:weightHistogram}
\end{figure}

In a nominal lifting scenario with a \SI{6}{\m} arm reach, the maximum rated capacity of the CASE250 is \SI{9210}{\kg} \cite{CASEdatasheet}, indicating a $3\sigma$ accuracy of approximately 1\% full-scale. 
With a lifting capacity of \SI{6000}{\kg}, our experiments confirm a ~1\% full-scale accuracy on M545.
\cref{fig:weightHistogram} shows an exemplary measurement distribution of 52 measurements with a true payload of \SI{580}{\kg}.
Measurements include the challenging conditions mentioned above that are found in real-world data.

Running on embedded hardware, the optimization has an average runtime of \SI{0.4}{\s} with a maximum runtime of \SI{1.4}{\s}.
With an average digging cycle time of about \SI{30}{\s} \cite{10616135}, our technique is well suited for real-world deployment.

The quasistatic reference method fails to estimate the correct payload and is highly influenced by acceleration disturbances, as indicated by a high measurement standard deviation (Std.), reflecting noise in the signal.
The commercial solution shows a significant deviation from the ground truth while maintaining a relatively low std. 
Various poses and payload attachments were tested to confirm the independence of this error from bucket loading. 
Notably, the commercial solution did not update measurements during accelerations.

\begin{table}
\caption{Error and standard deviation of benchmarked systems.}
\label{tab:comparison}
\begin{tabular}{ |c|c|c|c| } 
\hline
 & \begin{tabular}{@{}c@{}}Ours\\ CASE / M545\end{tabular} & quasistatic & Commercial \\
\hline
Error & 5.7 / \SI{4.5}{\kg}  & \SI{264.7}{\kg} & \SI{101.7}{\kg}\\ 
Std &        30.4 / \SI{20.0}{\kg} & \SI{2263}{\kg} & \SI{51.3}{\kg}\\ 
\hline
\end{tabular}
\end{table}

\subsection{EE Force Estimation}
\begin{figure}
\centering
\includegraphics[width=\linewidth]{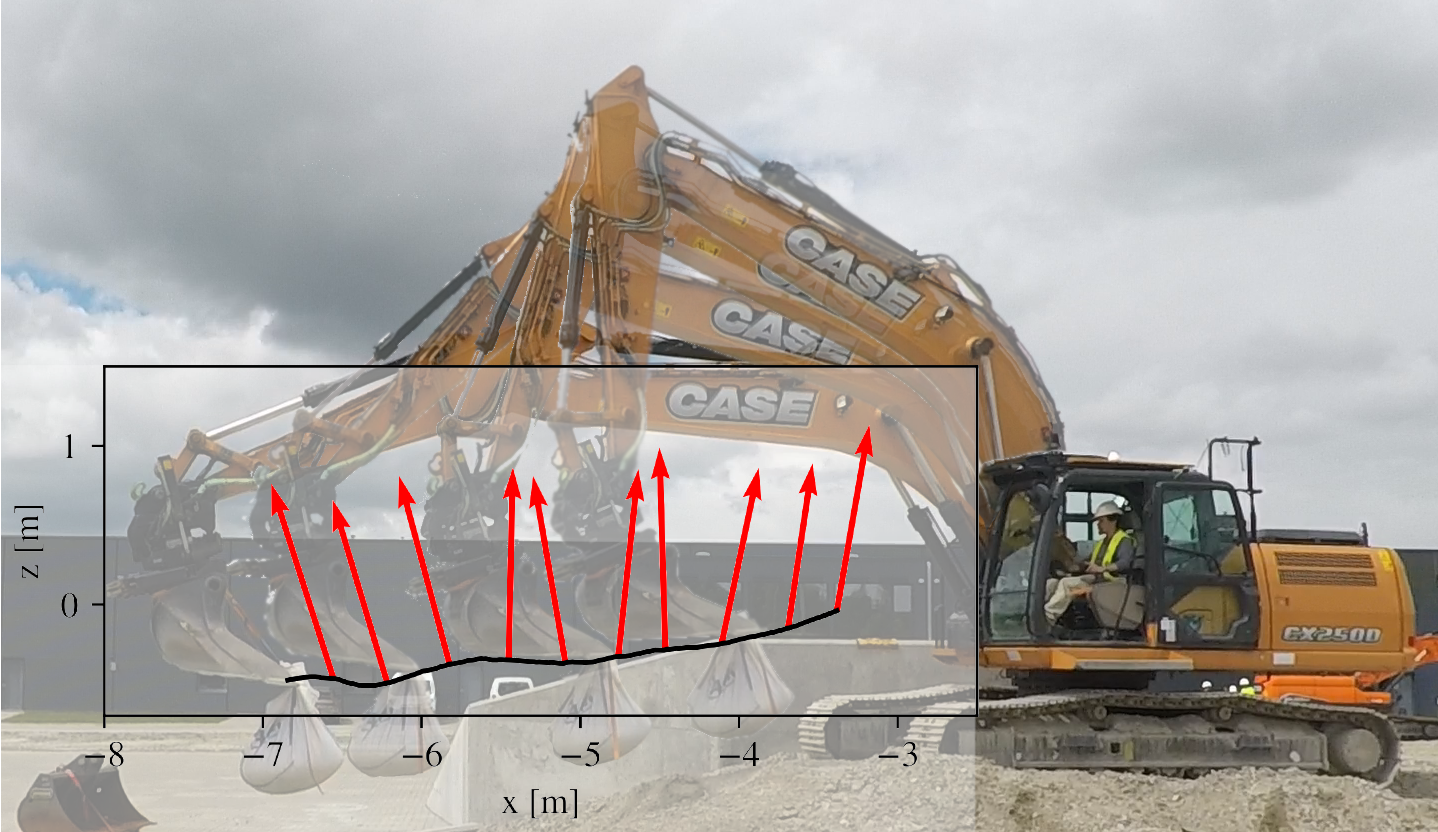}
\caption{Force vector estimation evaluation on CASE250}
\label{fig:comb}
\end{figure}

We evaluate the full 2D \ac{EE} force vector estimation by performing horizontal grading motions with a payload suspended from the shovel blade. 
This way, the undisturbed ground-truth force vector is aligned with gravity and has constant and known magnitude.
Oscillations of the suspended payload were inevitable and visible in the data, causing inflated estimated errors. 
\cref{fig:comb} shows an exemplary motion with estimated force vectors at \SI{1}{\s} intervals for plotting.
We evaluate the magnitude and direction of the estimated force vector on five grading motions of different speed and target height.

Metrics are calculated on all of the dense, \SI{50}{\Hz} force vector measurements.
The suspended payload of \SI{580.0}{\kg} (ground truth force: \SI{5689.8}{\N} aligned with gravity) was estimated on average as \ac{EE} force of \SI{6183}{\N} with a standard deviation of \SI{452}{\N}. 
The absolute angle error between the estimated force vector and the vertical gravity axis is \SI{13}{\deg} with a standard deviation of \SI{9.2}{\deg}.
Since the force vector calculation from measurements is $\mathcal{O}(1)$, its runtime is insignificant, independend on the executing hardware.

Qualitative evaluation of the \ac{EE} force vector estimation during soil interaction and digging shows that the estimated force vector direction matches the expected directions.

\subsection{Discussion}
The proposed method demonstrates high accuracy, even during challenging trajectories. 
The remaining errors in individual test cases can originate from several sources: movement of oil in and out of the cylinders changes the link \acp{COG} in an unmodeled way; small parameter estimation errors and simplifications in the friction model cause noticeable offsets; and joint acceleration measurements, relying on temporal differentiation of a noisy gyroscope, introduce additional uncertainty in estimated inertia torques. This last point is particularly relevant for highly dynamic motions.
Machine tilt deviations from the horizontal ground assumption are an additional source of error.

\section{Conclusion and future work}
\subsection{Conclusion}
\label{sec:conclusion}
This work presents a streamlined optimization-based method to estimate payload weight and \ac{EE} interaction forces on hydraulic excavators. 
The approach relies solely on kinematic joint measurements, cylinder pressures, and basic geometric measurements, thus avoiding the need for disassembly or specialized hardware. 
It uses lumped dynamic parameters to capture inertial, frictional, and gravitational effects calibrated through short identification sequences. 
Tests on two different excavator platforms show an accuracy of approximately one percent of the rated lifting capacity, which exceeds that of quasistatic methods and commercial solutions available. 
The estimator also provides real-time force measurements that can enable closed-loop control in tasks such as grading or sub-surface obstacle handling.

A key innovation is the unification of all dynamic effects into a model-based calibration pipeline that applies broadly to standard excavator configurations. 
Real-time execution of the estimator makes the force estimation ready for closed-loop control applications.
Model explainability and lightweight computations enable the presented method to be deployed on embedded devices as commonly found in industrial construction environments. 

\subsection{Future Work}
Using \ac{EE} force information is a necessity for machine agnostic controllers for earthworks.
Future work will include the development of \ac{EE} impedance control as well as digging controllers aware of sub-surface collisions and machine power constraints.
Precision grading in in inhomogeneous soil is an interesting task enabled by the presented work.
Moreover, heavy duty manipulation tasks such as demolition, material handling or rock stacking are tasks that will benefit from accurate force estimates.
Further improvements on the method itself might include work on a higher precision friction model as well as the inclusion of Coriolis forces in high-velocity multi-joint trajectories.
The use of the identified system dynamics in an observer has the potential of improving the \ac{EE} force estimates over the presented straight forward calculation from differential torques.
Automatically generated calibration trajectories would improve the consistency in accuracy of the estimated parameters and would also pose an interesting challenge for future investigations.
The use of a an estimated correction function for unmodeled dynamics could further enhance the estimators' precision.


\bibliography{sn-bibliography}

\end{document}